\documentclass[11pt,a4paper]{article}
\usepackage[hyperref]{acl2019}
\usepackage{times}
\usepackage{latexsym}
\usepackage{graphicx}
\usepackage{url}
\usepackage{tabularx}
\usepackage{makecell}
\usepackage{multirow}
\usepackage{enumitem}

\renewcommand\cellgape{\Gape[4pt]}
\definecolor{seagreen}{rgb}{0.0, 0.5, 0.0}
\definecolor{purple}{rgb}{0.47, 0.32, 0.66}
\aclfinalcopy 

\newlist{myitemize}{itemize}{3}

\title{Naver Labs Europe's Systems for the\\ Document-Level Generation and Translation Task at WNGT 2019}

\author{
	Fahimeh Saleh\thanks{\hspace{.15cm}This work was done while the author was visiting at Naver Labs Europe.} \\
	Monash University \\
	\texttt{fahimeh.saleh@monash.edu} \\\And
	Alexandre B\'erard \\
	\hspace{5.5cm} Naver Labs Europe \\
	\hspace{5.5cm}\texttt{first.last@naverlabs.com} \\
	\hspace{5.5cm}\url{https://europe.naverlabs.com} \\\And
	Ioan Calapodescu \\
	\\
	\\\AND
	Laurent Besacier \\
	Universit\'e Grenoble-Alpes \\
	\texttt{laurent.besacier@univ-grenoble-alpes.fr} \\  
}
\date{}

\usepackage[utf8]{inputenc}
\usepackage[T1]{fontenc}

\begin{document}
	\maketitle
	\begin{abstract}
		Recently, neural models led to significant improvements in both machine translation (MT) and natural language generation tasks (NLG).
		However, generation of long descriptive summaries conditioned on structured data remains an open challenge.
		Likewise, MT that goes beyond sentence-level context is still an open issue (e.g., document-level MT or MT with metadata). To address these challenges, we propose to leverage data from both tasks and do transfer learning between MT, NLG, and MT with source-side metadata (MT+NLG).
		First, we train document-based MT systems with large amounts of parallel data. Then, we adapt these models to pure NLG and MT+NLG tasks by fine-tuning with smaller amounts of domain-specific data.
		This end-to-end NLG approach, without data selection and planning, outperforms the previous state of the art on the Rotowire NLG task.
		We participated to the ``Document Generation and Translation'' task at WNGT 2019, and ranked first in all tracks.
	\end{abstract}
	
	\section{Introduction}
	
	Neural Machine Translation (NMT) and Neural Language Generation (NLG) are the top lines of the recent advances in Natural Language Processing.
	Although state-of-the-art NMT systems have reported impressive performance on several languages, there are still many challenges in this field especially when context is considered. Currently, the majority of NMT models translate sentences independently, without access to a larger context (e.g., other sentences from the same document or structured information). Additionally, despite improvements in text generation, generating long descriptive summaries conditioned on structured data is still an open challenge (e.g., table records). Existing models lack accuracy, coherence, or adequacy to source material \cite{wiseman2017challenges}.
	
	The two aspects which are mostly addressed in data-to-text generation techniques are identifying the most important information from input data, and verbalizing data as a coherent document: \textit{``What to talk about and how?"} \cite{mei2016talk}. These two challenges have been addressed separately as different modules in pipeline systems~\cite{McKeown1985,reiter2000building} or in an end-to-end manner with PCFGs or SMT-like approaches~\cite{wong2007,angeli2010,konstas2013}, or more recently, with neural generation models~\cite{wiseman2017challenges,lebret2016neural,mei2016talk}.
	In spite of generating fluent text, end-to-end neural generation models perform weakly in terms of best content selection~\cite{wiseman2017challenges}.
	Recently, \citet{puduppully2019data} trained an end-to-end data-to-document generation model on the Rotowire dataset (English summaries of basketball games with structured data).\footnote{\url{https://github.com/harvardnlp/boxscore-data}} They aimed to overcome the shortcomings of end-to-end neural NLG models by explicitly modeling content selection and planning in their architecture.
	
	We suggest in this paper to leverage the data from both MT and NLG tasks with transfer learning. As both tasks have the same target (e.g., English-language stories), they can share the same decoder. 
	The same encoder can also be used for NLG and MT if the NLG metadata is encoded as a text sequence.
	We first train domain-adapted document-level NMT models on large amounts of parallel data. Then we fine-tune these models on small amounts of NLG data, transitioning from MT to  NLG.
	We show that separate data selection and ordering steps are not necessary if  NLG model is trained at  document level and is given enough information. We propose a compact way to encode the data available in the original database, and enrich it with some extra facts that can be easily inferred with a minimal knowledge of the task. We also show that  NLG models trained with this data  capture document-level structure and can select and order information by themselves.
	
	\newcolumntype{H}{>{\setbox0=\hbox\bgroup}c<{\egroup}@{}}
	
	\section{Document-Level Generation and Translation Task}
	The goal of the Document-Level Generation and Translation (DGT) task  is to generate summaries of basketball games,
	in two languages (English and German),
	by using either structured data about the game,
	a game summary in the other language,
	or a combination of both.
	The task features 3 tracks, times 2 target languages (English or German):
	\textbf{NLG} (Data to Text),
	\textbf{MT} (Text to Text)
	and \textbf{MT+NLG} (Text~+~Data to Text).
	The data and evaluation are document-level,
	encouraging participants to generate full documents,
	rather than sentence-based outputs.
	Table \ref{tab:data} describes the allowed parallel and monolingual corpora.
	
	\begin{table}
		\centering
		\small
		\begin{tabular}{l|c|c|c|cH}
			Corpus & Lang(s) & Split & Docs & Sents & Words \\
			\hline
			\multirow{3}{*}{DGT} & \multirow{3}{*}{EN-DE} & train & 242 & 3247 & 78k \\
			& & valid & 240 & 3321 & 79k \\
			& & test & 241 & 3248 & 79k \\
			\hline
			\multirow{3}{*}{Rotowire} & \multirow{3}{*}{EN} & train & 3398 & 45.5k & 1.1M \\
			& & valid & 727 & 9.9k & 236k \\
			& & test & 728 & 10.0k & 241k \\
			\hline
			WMT19-sent & \multirow{2}{*}{EN-DE} & \multirow{2}{*}{train} & -- & 28.5M & 569M \\
			WMT19-doc & & & 68.4k & 3.63M & 88.5M \\
			\hline
			\multirow{2}{*}{News-crawl} & EN & \multirow{2}{*}{train} & 14.6M & 420M & 9.27B \\
			& DE & & 25.1M & 534M & 9.45B \\
		\end{tabular}
		\caption{Statistics of the allowed resources. The English sides of DGT-train, valid and test are respectively subsets of Rotowire-train, valid and test. More monolingual data is available, but we only used Rotowire and News-crawl. }
		\label{tab:data}
	\end{table}
	
	\section{Our MT and NLG Approaches}
	
	All our models (MT, NLG, MT+NLG) are based on Transformer Big \cite{vaswani2017attention}. Details for each track are given in the following sections.
	
	\subsection{Machine Translation Track}\label{sect:MT}
	
	For the MT track, we followed these steps:
	\vspace{-.1cm}
	\begin{enumerate}[leftmargin=.5cm]
		\itemsep0em
		\item Train sent-level MT models on all the WMT19 parallel data (doc and sent) plus DGT-train.
		\item Back-translate (BT) the German and English News-crawl by sampling \cite{edunov_understanding_2018}.
		\item Re-train sentence-level MT models on a concatenation of the WMT19 parallel data, DGT-train and BT. The later was split into 20 parts, one part for each training epoch. This is almost equivalent to oversampling the non-BT data by 20 and doing a single epoch of training.
		\item Fine-tune the best sentence-level checkpoint (according to valid perplexity) on document-level data. Like \citet{junczys2019microsoft}, we truncated the WMT documents into sequences of maximum 1100 BPE tokens. We also aggregated random sentences from WMT-sent into documents, and upsampled the DGT-train data. Contrary to \citet{junczys2019microsoft}, we do not use any sentence separator or document boundary tags.
		\item Fine-tune the best doc-level checkpoint on DGT-train plus back-translated Rotowire-train and Rotowire-valid.
	\end{enumerate}
	
	We describe the pre-processing and hyperparameters in Section~\ref{ssec:experiment}. In steps (1) and (3), we train for at most 20 epochs, with early stopping based on newstest2014 perplexity. In step (4), we train for at most 5 additional epochs, with early stopping according to DGT-valid perplexity (doc-level). In the last step, we train for 100 epochs, with BLEU evaluation on DGT-valid every 10 epochs. We also compute the BLEU score of the best checkpoint according to DGT-valid perplexity, and keep the checkpoint with highest BLEU.
	
	The models in step (5) overfit very quickly, reaching their best valid perplexity after only 1 or 2 epochs. For DE-EN, we found that the best DGT-valid BLEU was achieved anywhere between 10 and 100 epochs (sometimes with a high valid perplexity). For EN-DE, perplexity and BLEU correlated better, and the best checkpoint according to both scores was generally the same. The same observations apply when fine-tuning on NLG or MT+NLG data in the next sections.
	
	Like \citet{berard:2019:WMT}, all our MT models use corpus tags: each source sentence starts with a special token which identifies the corpus it comes from (e.g., Paracrawl, Rotowire, News-crawl). At test time, we use the DGT tag.
	
	One thing to note, is that document-level decoding is much slower than its sentence-level counterpart.\footnote{On a single V100, sent-level DGT-valid takes 1 minute to translate, while doc-level DGT-valid takes 6 minutes.} The goal of this document-level fine-tuning was not to increase translation quality, but to allow us to use the same model for MT and NLG, which is easier to do at the document level.
	
	\subsection{Natural Language Generation Track}\label{sect:NLG}
	Original metadata consists of one JSON document per game, containing information about teams and their players. We first generate compact representations of this metadata as text sequences. Then, we fine-tune our doc-level MT models (from step 4) on the NLG task by using this representation on the source side and full stories on the target side. We train on a concatenation of DGT-train, Rotowire-train and Rotowire-valid. We filter the later to remove games that are also in DGT-valid. Our metadata has the following structure:
	\begin{enumerate}[leftmargin=.5cm]
		\itemsep0em
		\small
		\item Date of the game as text.
		\item Home team information (winner/loser tag, team name and city, points in the game, season wins and losses and team-level scores) and information about its next game (date, home/visitor tag, other team's name), inferred from the other JSON documents in Rotowire-train.
		\item Visiting team information and details on its next game.
		\item $N$ best players of the home team (player name, followed by all his non-zero scores in a fixed order and his starting position). Players are sorted by points first, then by rebounds and assists.
		\item $N$ best players of the visiting team.
	\end{enumerate}
	To help the models identify useful information, we use a combination of special tokens and positional information. For instance, the home team is always first, but a \texttt{<WINNER>} tag precedes the winning team and its players. We ignore all-zero statistics, but always use the same position for each type of score (e.g., points, then rebounds, then assists) and special tokens to help identify them (e.g., \texttt{<PTS> 16} and \texttt{<REB> 8}). We try to limit the number of tags to keep the sequences short (e.g., made and attempted free throws and percentage: \texttt{<FT> 3 5 60}). An example of metadata representation is shown in Table~\ref{tab:example-data-compressed}. 
	
	\begin{table*}[t!]
		\centering
		\tiny
		\begin{tabular}{|c|p{14cm}|}
			\hline
			Metadata & \texttt{<DATE> Freitag Februar 2017 <WINNER> Oklahoma City Thunder <PTS> 114 <WINS> 29 <LOSSES> 22 <REB> 47 <AST> 21 <TO> 20 <FG> 38 80 48 <FG3> 13 26 50 <FT> 25 33 76 <NEXT> Sonntag Februar 2017 <HOME> Portland Trail Blazers <LOSER> Memphis Grizzlies <PTS> 102 <WINS> 30 <LOSSES> 22 <REB> 29 <AST> 21 <TO> 12 <FG> 40 83 48 <FG3> 3 19 16 <FT> 19 22 86 <NEXT> Samstag Februar 2017 <VIS> Minnesota Timberwolves <WINNER> <PLAYER> Russell Westbrook <PTS> 38 <REB> 13 <AST> 12 <STL> 3 <PF> 2 <FG> 8 20 40 <FG3> 5 7 71 <FT> 17 17 100 <POS> Guard <PLAYER> Steven Adams <PTS> 16 <REB> 12 <AST> 2 <STL> 1 <BLK> 2 <PF> 4 <FG> 7 13 54 <FT> 2 6 33 <POS> Center <PLAYER> Joffrey Lauvergne <PTS> 16 <REB> 8 <AST> 2 <PF> 3 <FG> 6 7 86 <FG3> 3 4 75 <FT> 1 2 50 <POS> Bank <LOSER> <PLAYER> Marc Gasol <PTS> 31 <REB> 4 <AST> 8 <STL> 2 <BLK> 1 <PF> 4 <FG> 14 24 58 <FG3> 0 4 0 <FT> 3 3 100 <POS> Center <PLAYER> Mike Conley <PTS> 18 <REB> 1 <AST> 2 <STL> 3 <FG> 7 16 44 <FG3> 1 5 20 <FT> 3 5 60 <POS> Guard <PLAYER> Zach Randolph <PTS> 16 <REB> 10 <AST> 3 <STL> 1 <PF> 4 <FG> 6 14 43 <FG3> 0 1 0 <FT> 4 4 100 <POS> Bank
			}
			\\
			\hline
			Reference story & \textcolor{seagreen}{The Oklahoma City Thunder defeated the visiting Memphis Grizzlies 114 - 102 , \textcolor{blue}{at Chesapeake Energy Arena} on Friday} evening~. The Grizzlies led by four after three quarters , but then \textcolor{seagreen}{Russell Westbrook} went absolutely ballistic in the fourth quarter , \textcolor{seagreen}{scoring 19 points} in the quarter , \textcolor{blue}{including 15 points straight and unanswered} , \textcolor{seagreen}{to take his team from down \textcolor{blue}{102 - 99} to the final score of \textcolor{seagreen}{114 - 102}}~. This snaps the Grizzlies \textcolor{blue}{three-game win streak} , while \textcolor{seagreen}{Westbrook} added to his ridiculous \textcolor{seagreen}{\textcolor{blue}{ triple-double}} count , as \textcolor{blue}{he notched his 25th of the season}~. \textcolor{seagreen}{The Thunder ( 29 - 22 )} only scored \textcolor{blue}{21 points in the first quarter} , before \textcolor{seagreen}{\textcolor{blue}{outscoring the Grizz by 12}} in the second , to take an \textcolor{blue}{eight-point lead into half time}~. They were then \textcolor{blue}{outscored by 12 in the third} , and \textcolor{blue}{entered the fourth down by four}~. \textcolor{blue}{The Thunder outscored the Grizz by 14 in the fourth} , behind Russell Westbrook and his takeover~.\textcolor{seagreen}{Westbrook finished with 38 points , 13 rebounds , 12 assists and three steals to lead his squad to a big win}~. Westbrook also matched a \textcolor{blue}{career-best} as he went \textcolor{seagreen}{17-of-17 from the foul line}~. \textcolor{blue}{Three other players scored in double digits for the Thunder} , including \textcolor{seagreen}{Steven Adams} and \textcolor{seagreen}{Joffrey Lauvergne} , who pitched in \textcolor{seagreen}{16 points} each~. \textcolor{seagreen}{The Thunder go on to visit Portland on Sunday~.} Meanwhile , \textcolor{seagreen}{the Grizzlies ( 30 - 22 )} saw their \textcolor{blue}{win streak come to an end} as \textcolor{blue}{they went scoreless in the last 2:50} as they were stifled by an overwhelming push of energy and intensity from the Thunder and their home crowd~. \textcolor{seagreen}{Marc Gasol} finished with a team-high \textcolor{seagreen}{31 points , to go along with eight assists , four rebounds , two steals and one block} , in 42 minutes~. \textcolor{seagreen}{Zach Randolph} came up big in 29 minutes \textcolor{seagreen}{off the bench} , with \textcolor{seagreen}{16 points and 10 rebounds} , while \textcolor{seagreen}{Mike Conley chipped in 18 points , three steals and two assists }. \textcolor{seagreen}{Chandler Parsons added 12 points on 3-of-6 shooting} , in 25 minutes~. \textcolor{seagreen}{The Grizzlies will have to put this one in the past , as they go right back in action on Saturday , when they visit Minnesota to take on the Timberwolves~.}
			\\
			\hline
			Generated story & \textcolor{seagreen}{The Oklahoma City Thunder defeated the Memphis Grizzlies~, 114 - 102~, at \textcolor{blue}{Chesapeake Energy Arena} on Friday .} Despite there being just \textcolor{red}{five ties and six lead} changes in this game~, it was close throughout . In fact~, \textcolor{blue}{neither team led by more than 12 points}~, as the \textcolor{red}{nine-point final margin} was one of the largest leads of the game . \textcolor{seagreen}{Oklahoma City actually won this game}~, \textcolor{blue}{despite being out shot from the field~, from three-point range and from the free-throw line} . \textcolor{blue}{Rebounding was the decisive statistic~, with the Thunder winning that battle~,} \textcolor{seagreen}{47 - 29} . \textcolor{red}{They also dominated the assist-to-turnover ratio~, recording 10 more assists and committing three less turnovers} . \textcolor{seagreen}{The Grizzlies ( 30 - 22 )} have now \textcolor{red}{lost three of their last four games}~, as they continue to fall down the standings . \textcolor{seagreen}{Marc Gasol did all he could for Memphis~, as he led the team with 31 points~, four rebounds~, eight assists and two steals} . \textcolor{blue}{Mike Conley finished second on the team}~, \textcolor{seagreen}{totaling 18 points~, two assists and three steals} . \textcolor{orange}{Mike Conley finished second on the team~, amassing 18 points and three steals}~. \textcolor{seagreen}{Zach Randolph was huge off the bench~, providing 16 points~, 10 rebounds and three assists} . \textcolor{seagreen}{The Grizzlies will be off until Saturday~, when they play the Minnesota Timberwolves on the road for their next game} . \textcolor{seagreen}{ The Thunder ( 29~-~22 )} have rode \textcolor{blue}{Russell Westbrook 's triple-double} train all year and they got another one here . \textcolor{seagreen}{Westbrook collected 38 points~, 13 rebounds~, 12 assists and three steals}~. \textcolor{seagreen}{Steven Adams \textcolor{blue}{recorded a double-double}~, amassing 16 points and 12 rebounds . Joffrey Lauvergne was a nice spark off the bench~, providing 16 points and eight rebounds . The Thunder will look to keep rolling on Sunday against the Portland Trail Blazers .}
			\\
			\hline
		\end{tabular}
		\vspace{-.1cm}
		\caption{\textbf{Metadata:} our metadata encoding. 
			\textbf{Reference story:} story \#48 from DGT-valid.
			\textbf{Generated story:} output of the English NLG model (3-player).
			\textbf{Green}: text based on facts from the metadata. 
			\textbf{Blue:} correct facts which are not explicitly in the metadata.
			\textbf{Red}: hallucinations or incorrect facts. 
			\textbf{Orange}: repetitions.} \label{tab:example-data-compressed}
	\end{table*}
	
	\subsection{MT+NLG Track}\label{sect:TGMT}
	
	For the MT+NLG track, we concatenate the MT source with the NLG data. We use the same metadata encoding method as in the NLG track and we fine-tune our doc-level MT models (from step 4).
	We also randomly mask tokens in the MT source (by replacing them with a \texttt{<MASK>} token), with 20\% or 50\% chance (with one different sampling per epoch). The goal is to force the model to use the metadata because of missing information in the source. At test time, we do not mask any token.
	
	\section{Experiments}
	\label{ssec:experiment}
	\subsection{Data Pre-processing}
	We filter the WMT19-sent parallel corpus with \texttt{langid.py} \cite{lui_2012} and remove sentences of more than 175 tokens or with a length ratio greater than 1.5. Then, we apply the official DGT tokenizer (based on NLTK's \texttt{word\_tokenize}) to the non-tokenized text (everything but DGT and Rotowire).
	
	We apply BPE segmentation \cite{sennrich_neural_2016} with a joined SentencePiece-like model \cite{kudo_sentencepiece:_2018}, with 32k merge operations, obtained on WMT + DGT-train (English~+~German). The vocabulary threshold is set to 100 and inline casing is applied \cite{berard:2019:WMT}. We employ the same joined BPE model and Fairseq dictionary for all models. The metadata is translated into the source language of the MT model used for initialization,\footnote{Only week days, months and player positions need to be translated.} and segmented into BPE (except for the special tokens) to allow transfer between MT and NLG. Then, we add a corpus tag to each source sequence, which specifies its origin (Rotowire, News-crawl, etc.)
	
	Like \citet{junczys2019microsoft}, we split WMT19 documents that are too long into shorter documents (maximum 1100 BPE tokens). We also transform the sent-level WMT19 data into doc-level data by shuffling the corpus and grouping consecutive sentences into documents of random length. Finally, we upsample the doc-level data (WMT19 and DGT) by 8 times its original size (in terms of sent count). We do so by sampling random spans of consecutive sentences until reaching the desired size.
	
	The DGT and Rotowire data is already tokenized and does not need filtering nor truncating. We segment it into BPE units and add corpus tags.
	
	\subsection{Settings}
	
	All the models are Transformer Big \cite{vaswani2017attention}, implemented in Fairseq \cite{ott_scaling_2018}.
	We use the same hyper-parameters as \citet{ott_scaling_2018}, with Adam and an inverse square root schedule with warmup (maximum LR 0.0005).
	We apply dropout and label smoothing with a rate of 0.1. The source and target embeddings are shared and tied with the last layer. We train with half-precision floats on 8 V100 GPUs, with at most 3500 tokens per batch and delayed updates of 10 batches.
	When fine-tuning on DGT-train or Rotowire + DGT-train (Step 5 of the MT track, or NLG/MT+NLG fine-tuning), we use a fixed learning rate schedule (Adam with 0.00005 LR) and a much smaller batch size (1500 tokens on a single GPU without delayed updates). We train for 100 epochs, compute DGT-valid perplexity at each epoch, and DGT-valid BLEU every 10 epochs.
	
	\subsection{BLEU evaluation}
	
	\begin{table}
		\centering
		\small
		\begin{tabular}{l|c|c|c|c}
			Track & Target & Constrained & Valid & Test\\\hline
			NLG & \multirow{4}{*}{EN} & no & 23.5 & 20.5 \\ 
			MT & & yes & 60.2 & 58.2 \\ 
			MT & & no & 64.2 & 62.2 \\ 
			MT+NLG & & yes & 64.4 & 62.2 \\\hline 
			NLG & \multirow{3}{*}{DE} & no & 16.9 & 16.1 \\ 
			MT & & yes & 49.8 & 48.0 \\ 
			MT+NLG & & yes & 49.4 & 48.2 \\ 
		\end{tabular}
		\caption{Doc-level BLEU scores on the DGT valid and test sets of our submitted models in all tracks.}
		\label{tab:BLEU_all_tracks}
	\end{table}
	
	\paragraph{Submitted models.} For each track, we selected the best models according to their BLEU score on DGT-valid. The scores are shown in Table \ref{tab:BLEU_all_tracks}, and a description of the submitted models is given in Table \ref{tab:TracksDetails}. We compute BLEU using SacreBLEU with its tokenization set to \emph{none},\footnote{SacreBLEU signature: \emph{BLEU+case.mixed+numrefs.1+\\smooth.exp+tok.none+version.1.3.1}} as the model outputs and references are already tokenized with NLTK. \citet{Hayashi2019} give the full results of the task: the scores of the other participants, and values of other metrics (e.g., ROUGE).
	Our NLG models are ``unconstrained'' because the WMT19 parallel data, which we used for pre-training, was not allowed in this track.
	Similarly, we do two submissions for DE-EN MT: one constrained, where we fine-tuned the doc-level MT model on DGT-train only, and one unconstrained, where we also used back-translated Rotowire-train and valid. All the MT and MT+NLG models are ensembles of 5 fine-tuning runs.
	Cascading the English NLG model with the ensemble of EN-DE MT models gives a BLEU score of 14.9 on DGT-test, slightly lower than the end-to-end German NLG model (16.1).
	We see that in the same data conditions (unconstrained mode), the MT+NLG models are not better than the pure MT models. Furthermore, we evaluated the MT+NLG models with MT-only source, and found only a slight decrease of $\approx0.3$ BLEU, which confirms our suspicion that the NLG information is mostly ignored.
	
	\paragraph{NMT analysis.} Table~\ref{tab:MT_BLEU} shows the BLEU scores of our MT models at different stages of training (sent-level, doc-level, fine-tuned), and compares them against one of the top contestants of the WMT19 news translation task \cite{fair2019}.
	
	\renewcommand\cellgape{\Gape[1pt]}
	
	\begin{table}[t!]
		\centering
		\scriptsize
		\begin{tabular}{l|l|l}
			Track       & N best players & Details \\ \hline
			NLG (EN)    & 4              & Rotowire BT + DGT-train + tags \\ \hline
			NLG (DE)    & 6              & Rotowire BT + DGT-train + tags \\ \hline
			MT (DE-EN) & N/A & \makecell[l]{
				\textit{Unconstrained:} Rotowire BT + \\ DGT-train + tags + ensemble \\ \textit{Constrained:} DGT-train only + \\ ensemble
			} \\ \hline
			MT (EN-DE) & N/A & DGT-train only + ensemble \\ \hline
			MT+NLG (EN) & 3 & \makecell[l]{Rotowire BT + DGT-train + 20\% \\ text masking + tags + ensemble} \\ \hline
			MT+NLG (DE) & 3 & \makecell[l]{Rotowire BT + DGT-train + \\ tags + ensemble} \\
		\end{tabular}
		\vspace{-.2cm}
		\caption{Description of our submissions.}
		\label{tab:TracksDetails}
	\end{table}
	
	\begin{table}[t!]
		\centering
		\small
		\begin{tabular}{c|c|c|c|c}
			Model & Target & Valid & Test & News 2019 \\\hline
			FAIR 2019 & \multirow{4}{*}{EN} & 48.5 & 47.7 & \textbf{41.0} \\ 
			Sent-level & & 55.6 & 54.2 & 40.9 \\ 
			Doc-level & & 56.5 & 55.0 & 38.5 \\ 
			Fine-tuned & & \textbf{61.7} & \textbf{59.6} & 21.7 \\ 
			\hline
			FAIR 2019 & \multirow{4}{*}{DE} & 37.5 & 37.0 & 40.8 \\ 
			Sent-level & & 47.3 & 46.7 & \textbf{42.9} \\ 
			Doc-level & & \textbf{48.2} & \textbf{47.5} & 41.6 \\ 
			Fine-tuned & & 48.0 & 46.7 & 41.3 \\ 
		\end{tabular}
		\caption{BLEU scores of the MT models at different stages of training, and comparison with the state of the art. Scores on DGT-valid and DGT-test are doc-level, while News 2019 is sent-level (and so is decoding). On the latter, we used the DGT corpus tag for DE-EN, and the Paracrawl tag for EN-DE (we chose the tags with best BLEU on newstest2014). Scores by the ``fine-tuned'' models are averaged over 5 runs.}
		\label{tab:MT_BLEU}
	\end{table}
	
\begin{figure}[t!]
		\centering
		\includegraphics[width=6.5cm]{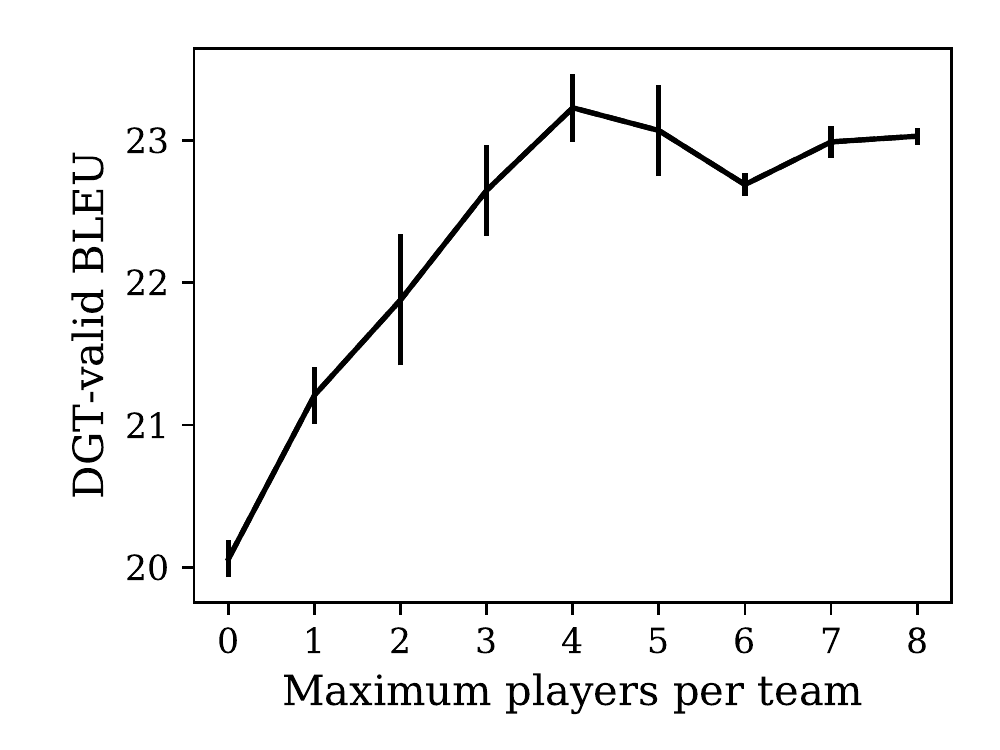}
		\vspace{-.3cm}
		\caption{DGT-valid BLEU (by the best checkpoint) depending on the maximum number of selected players for the English NLG track.}
		\label{fig:BLEU_vs_players}
	\end{figure}

	\paragraph{English NLG analysis.}
	Table~\ref{tab:NLG_SOTA} shows a 5.7 BLEU improvement on Rotowire-test by our English NLG model compared to the previous state of the art.
	Figure~\ref{fig:BLEU_vs_players} shows the DGT-valid BLEU scores of our English NLG models when varying the number of players selected in the metadata. We see that there is a sweet spot at 4, but surprisingly, increasing the number of players up to 8 does not degrade BLEU significantly. We hypothesize that because the players are sorted from best to worst, the models learn to ignore the last players.
	
	\begin{table}
		\centering
		\begin{tabular}{c|c}
			Model & Rotowire test \\
			\hline
			\citet{wiseman2017challenges} & 14.5 \\
			\citet{puduppully2019data} & 16.5 \\
			Ours (4-player) & 22.2 \\
		\end{tabular}
		\vspace{-.1cm}
		\caption{English NLG comparison against state-of-the-art on Rotowire-test. BLEU of submitted NLG (EN) model, averaged over 3 runs. Because Rotowire tokenization is slightly different, we apply a set of fixes to the model outputs (e.g., \texttt{1-of-3} $\to$ \texttt{1 - of - 3}).}
		\label{tab:NLG_SOTA}
	\end{table}
	
	\begin{table}[t!]
		\centering
		\small
		\begin{tabular}{l|c|c}
			Model & Valid & Test \\
			\hline
			Baseline (3 players, sorted) & \textbf{22.7} & 20.4 \\ 
			\hline
			No player & 20.1 & 18.8 \\ 
			All players, sorted & \textbf{22.7} & 20.9 \\ 
			All players, shuffled & 22.0 & 20.0 \\ 
			\hline
			(1) No next game & 22.0 & 19.9 \\ 
			(2) No week day & 22.2 & 20.5 \\ 
			(3) No player position & 22.6 & 20.5 \\ 
			(4) No team-level sums & 22.5 & 20.5 \\ 
			(5) Remove most tags & 22.6 & 20.8 \\ 
			\hline
			(1) to (5) & 21.3 & 19.7 \\ 
		\end{tabular}
		\vspace{-.1cm}
		\caption{English NLG ablation study, starting from a 3 best player baseline (the submitted NLG model has 4 players). BLEU averages over 3 runs. Standard deviation ranges between 0.1 and 0.4.}
		\label{tab:NLG_ablation_study}
	\end{table}
	
	From Table~\ref{tab:NLG_ablation_study}, we see that sorting players helps, but only slightly. Using only team-level information, and no information about players gives worse but still decent BLEU scores.
	
	Week day, player position or team-level aggregated scores can be removed without hurting BLEU. However, information about next games seems useful. 
	Interestingly, relying on position only and removing most tags (e.g., \texttt{<PTS>}, \texttt{<FT>}) seems to be fine. In this case, we also print all-zero stats, for the position of each statistic to be consistent across players and games.
	
	\begin{table*}[t]
		\small
		\centering
		\begin{tabular}{|c|c|}
			\hline
			Stadium name (+) &
			\makecell[l]{\textcolor{blue}{REF:} The Golden State Warriors ( 56 - 6 ) defeated the Orlando Magic ( 27 - 35 ) 119 - 113 \\ at \textbf{Oracle Arena} on Monday~.\\
				\textcolor{blue}{NLG:} The Golden State Warriors ( 56 - 6 ) defeated the Orlando Magic ( 27 - 35 ) 119 - 113 \\ on Monday at \textbf{Oracle Arena}~.} \\
			\hline
			Team alias (+) &
			\makecell[l]{\textcolor{blue}{REF:} The Heat held the \textbf{Sixers} to 38 percent shooting and blocked 14 shots in the win~. \\
				\textcolor{blue}{NLG:} The \textbf{Sixers} shot just 38 percent from the field and 32 percent from the three-point line , \\ while the Heat shot 44 percent from the floor and a meager 28 percent from deep~.} \\
			\hline
			\makecell[c]{Double-doubles or \\ triple-doubles (+)} &
			\makecell[l]{\textcolor{blue}{REF:} \textbf{Kevin Love 's 29-point , 13-rebound double-double} led the way for the Cavs , who 'd \\ rested Kyrie Irving on Tuesday~. \\
				\textcolor{blue}{NLG:} \textbf{Love} led the way for Cleveland with a \textbf{29-point , 13-rebound double-double} that also \\ included three assists and two steals~.}\\
			\hline
			Player injuries (-) &
			\makecell[l]{\textcolor{blue}{NLG:} The Timberwolves ( 28 - 44 ) checked in to Saturday 's contest with an injury-riddled \\ frontcourt , as Ricky Rubio \textbf{( knee )} and Karl-Anthony Towns \textbf{( ankle )} were sidelined~.} \\
			\hline
			Ranking (-) &
			\makecell[l]{\textcolor{blue}{NLG:} The Heat ( 10 - 22 ) fell to 10 - 22 and \textbf{remain in last place} in the Eastern Conference 's \\ Southeast Division~.} \\
			\hline
			\makecell[c]{Season-level \\ player stats (-)} &
			\makecell[l]{\textcolor{blue}{NLG:} It was a season-high in points for Thomas , \textbf{who 's now averaging 17 points per game} \\ \textbf{on the season}} \\
			\hline
		\end{tabular}
		\caption{Correctly predicted information that is not explicitly in the metadata (+), or hallucinations (-).}
		\label{tab:Qualitative_good}
	\end{table*}
	
	\vspace{-.1cm}
	\paragraph{Train-test overlap on Rotowire.} We found a significant overlap between Rotowire train and test: 222 out of 728 Rotowire-test games are also in Rotowire-train (68/241 for DGT-test). The corresponding stories are always different but bear many similarities (some sentences are completely identical). Rotowire-train gets 24.2 BLEU when evaluated against Rotowire-test (subset of 222 stories). This gives us an estimate of human-level performance on this task. Our submitted NLG model gets 21.8 on the same subset.
	This overlap may cause an artificial increase in BLEU, that would unfairly favor overfitted models. Indeed, when filtering Rotowire-train to remove games that were also in DGT test, we found a slight decrease in BLEU (19.8 instead of 20.4).
	
	\subsection{Qualitative evaluation}
	\vspace{-.05cm}
	
	As shown in Table \ref{tab:example-data-compressed}, the NLG model (3-player) has several good properties besides coherent document-level generation and the ability to ``copy" metadata. It has learned generic information about the teams and players. As such, it can generate relevant information which is absent from  metadata (see Table \ref{tab:Qualitative_good}).
	For example, the model correctly predicts the name of the stadium where the game was played. This implies that it knows which team is hosting (this information is encoded implicitly by the position of the team in the data), and what is the stadium of this team's city (not in the metadata).
	Other facts that are absent from the metadata, and predicted correctly nonetheless, are team aliases (e.g., the \textit{Sixers}) and player nicknames (e.g., \textit{the Greek Freak}). The model can also generate other surface forms for the team names (e.g., \textit{the other Cavalier}).
	
	The NLG model can infer some information from the structured data, like double-digit scores, ``double-doubles'' (e.g., when a player has more than 10 points and 10 assists) and ``triple-doubles''. On the other hand, some numerical facts are inaccurate (e.g., score differences or comparisons).
	Some facts which are not present in the structured data, like player injuries, season-level player statistics, current ranking of a team, or timing information are hallucinated. We believe that most of these hallucinations could  be avoided by adding the missing facts to the structured data. More rarely, the model duplicates a piece of information.
	
	Another of its flaws is a poor generalization to new names (team, city or player). This can quickly be observed by replacing a team name by a fictional one in the metadata. In this case, the model almost always reverts to an existing team.
	This may be due to overfitting, as earlier checkpoints seem to handle unknown team names better, even though they give lower BLEU. This generalization property could be assessed by doing a new train/test split, that does not share the same teams.

	\vspace{-.05cm}
	
	\section{Conclusion}
	
	\vspace{-.05cm}
	
	We participated in the 3 tracks of the DGT task: MT, NLG and MT+NLG. Our systems rely heavily on transfer learning, from document-level MT (high-resource task) to document-level NLG (low-resource task). Our submitted systems ranked first in all tracks.
	
	For the MT track, the usual domain adaptation techniques performed well. The MT+NLG models did not show any significant improvement over pure MT. The MT models are already very good and probably do not need the extra context (which is generally encoded in the source-language summary already). Finally, our NLG models, bootstrapped from the MT models, do fluent and coherent text generation and are even able to infer some facts that are not explicitly encoded in the structured data.
	Some of their current limitations (mostly hallucinations) could be solved by adding extra information (e.g., injured players, current team rank, number of consecutive wins, etc.)
	
	Our aggressive fine-tuning allowed us to specialize MT models into NLG models, but it will be interesting to study whether a single model can solve both tasks at once (i.e., with multi-task learning), possibly in both languages.
	
	\clearpage
	\bibliography{paper}
	\bibliographystyle{acl_natbib}
	
\end{document}